\documentclass[letterpaper]{article} 
\usepackage{aaai24}  
\usepackage{times}  
\usepackage{helvet}  
\usepackage{courier}  
\usepackage[hyphens]{url}  
\usepackage{graphicx} 
\usepackage{natbib}  
\usepackage{caption} 
\usepackage{algorithm}
\usepackage{algorithmic}

\usepackage{newfloat}
\usepackage{listings}
\DeclareCaptionStyle{ruled}{labelfont=normalfont,labelsep=colon,strut=off} 
\floatstyle{ruled}
\newfloat{listing}{tb}{lst}{}
\floatname{listing}{Listing}
\usepackage{amsmath}
\usepackage{amssymb}
\usepackage{tabularx}
\usepackage{multirow}
\usepackage{booktabs}
\usepackage[capitalise]{cleveref}
\usepackage{siunitx}
\usepackage{environ}
\newcolumntype{Y}{>{\centering\arraybackslash}X}

\def\arxiv@on{T}

\usepackage{tikz}

\usetikzlibrary{
  arrows.meta,
  shadows,
  positioning,
  shapes.arrows,
  shapes.geometric,
  decorations.pathreplacing,
  backgrounds,
  calc,
}

\tikzset{
    every shadow/.style={
        shadow xshift=.3ex,
        shadow yshift=-.3ex
    },
    frame/.style={
        rectangle, draw,
        text width=2.7em,
        text centered,
        minimum height=1.5em,
        fill=white,
        drop shadow,
        rounded corners=2,
    },
    big_frame/.style={
        rectangle, draw,
        text width=4.8em,
        text centered,
        minimum height=1.5em,
        fill=white,
        drop shadow,
        rounded corners=2,
    },
    line/.style={
        draw,
        line width=0.1pt,
        -{Latex},
    },
}

\newcommand{\temp}{\tau}
\newcommand{\epoch}{m}
\newcommand{\BigPi}{\Pi}

\def\bbbr{{\rm I\!R}} 

\title{Peer Learning: Learning Complex Policies in Groups from Scratch via Action Recommendations}
\author {
    Cedric Derstroff\textsuperscript{\rm 1,2},
    Mattia Cerrato\textsuperscript{\rm 3},
    Jannis Brugger\textsuperscript{\rm 1,2},
    Jan Peters\textsuperscript{\rm 1,2,4,5},
    Stefan Kramer\textsuperscript{\rm 3}
}
\affiliations {
    \textsuperscript{\rm 1} Technische Universität Darmstadt\\
    \textsuperscript{\rm 2} Hessian Center for Artificial Intelligence (hessian.AI)\\
    \textsuperscript{\rm 3} Johannes Gutenberg-Universität Mainz\\
    \textsuperscript{\rm 4} German Research Center for AI (DFKI)\\
    \textsuperscript{\rm 5} Centre for Cognitive Science\\
    \{cedric.derstroff, jannis.brugger, jan.peters\}@tu-darmstadt.de,
    mcerrato@uni-mainz.de,
    kramer@informatik.uni-mainz.de
}

\begin{document}

\maketitle

\begin{abstract}
\emph{Peer learning} is a novel high-level reinforcement learning framework for agents learning in groups. 
While standard reinforcement learning trains an individual agent in trial-and-error fashion, all on its own, peer learning addresses a related setting in which a group of agents, i.e., \textit{peers}, learns to master a task simultaneously together from scratch. Peers are allowed to communicate only about their own states and actions recommended by others: ``What would you do in my situation?''.
Our motivation is to study the learning behavior of these agents.
We formalize the teacher selection process in the action advice setting as a multi-armed bandit problem and therefore highlight the need for exploration. 
Eventually, we analyze the learning behavior of the peers and observe their ability to rank the agents' performance within the study group and understand which agents give reliable advice. Further, we compare peer learning with single agent learning and a state-of-the-art action advice baseline.
We show that peer learning is able to outperform single-agent learning and the baseline in several challenging discrete and continuous OpenAI Gym domains. Doing so, we also show that within such a framework 
complex policies from action recommendations beyond discrete action spaces can evolve.
\end{abstract}
%
%
\section{Introduction}
\label{sec:intro}
Learning by trial-and-error is one of the most influential theories of learning. 
Its roots go deeper than theories of \emph{human} learning, in ethology and particularly in Morgan's observations on animal behavior \cite{thorpe1979origins}.
Edward Thorndike's ``Law of Effect'', for instance, posits that animals select responses (\emph{actions}) to a given situation or stimulus (\emph{state}) based on the satisfaction that followed in previous occurrences of the same situation and action \cite{thorndike1898animal}.
The Law of Effect is the basis for the behaviorist view of biological learning as pioneered by Skinner \cite{skinner1988selection}, who introduced the term ``reinforcement'' to describe the conditioning effect of certain operants (responses, or \emph{rewards}) on the behavior of lab mice.
Computational investigations of behaviorism closely followed, from both Turing \cite{turing1951intelligent} and Minsky, who explicitly referenced the work done by Skinner in studying animal behavior as an influence to early artificial intelligence (AI) \cite{minsky1961steps}.

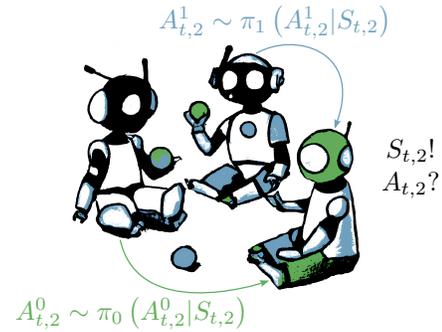
\begin{figure}[ht!]
\centering
\begingroup%
  \makeatletter%
  \providecommand\color[2][]{%
    \errmessage{(Inkscape) Color is used for the text in Inkscape, but the package 'color.sty' is not loaded}%
    \renewcommand\color[2][]{}%
  }%
  \providecommand\transparent[1]{%
    \errmessage{(Inkscape) Transparency is used (non-zero) for the text in Inkscape, but the package 'transparent.sty' is not loaded}%
    \renewcommand\transparent[1]{}%
  }%
  \providecommand\rotatebox[2]{#2}%
  \newcommand*\fsize{\dimexpr\f@size pt\relax}%
  \newcommand*\lineheight[1]{\fontsize{\fsize}{#1\fsize}\selectfont}%
  \ifx\svgwidth\undefined%
    \setlength{\unitlength}{166.31999588bp}%
    \ifx\svgscale\undefined%
      \relax%
    \else%
      \setlength{\unitlength}{\unitlength * \real{\svgscale}}%
    \fi%
  \else%
    \setlength{\unitlength}{\svgwidth}%
  \fi%
  \global\let\svgwidth\undefined%
  \global\let\svgscale\undefined%
  \makeatother%
  \begin{picture}(1,0.80801772)%
    \lineheight{1}%
    \setlength\tabcolsep{0pt}%
    \put(0,0){\includegraphics[width=\unitlength,page=1]{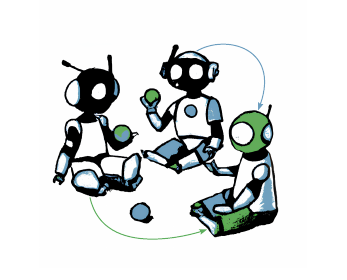}}%
    \put(0.91803989,0.42489068){\makebox(0,0)[t]{\lineheight{1.25}\smash{\begin{tabular}[t]{c}$S_{t,2}!$\\$A_{t,2}?$\end{tabular}}}}%
    \put(0.02241755,0.05716982){\color[rgb]{0.38823529,0.67843137,0.35294118}\makebox(0,0)[lt]{\smash{\begin{tabular}[t]{l}$A_{t,2}^0 \sim \pi_0 \left(A_{t,2}^0 | S_{t,2} \right)$\end{tabular}}}}%
    \put(0.34711348,0.71745561){\color[rgb]{0.43921569,0.62745098,0.74901961}\makebox(0,0)[lt]{\smash{\begin{tabular}[t]{l}$A_{t,2}^1 \sim \pi_1 \left(A_{t,2}^1 | S_{t,2} \right)$\end{tabular}}}}%
  \end{picture}%
\endgroup%
\caption{Peer learning: Agent $2$ (right) asks for advice about which action to perform at the current step $S_{t, 2}$ it finds itself in, and receives action suggestions $A^{0}_{t, 2}$ and $A^{1}_{t, 2}$ from agents $0$ and $1$, on the left and center,  respectively. The agents operate in their own environment and on their own complex (motor) task,  here: learning how to handle a ball.}
\label{fig:peer_learning}
\end{figure}
Learning as trial-and-error under a reinforcement signal has since been the core idea of reinforcement learning (RL) as a subfield of AI and machine learning (ML), as discussed by Sutton and Barto \cite{Sutton2018}.
Another classic line of investigation in learning theory seeks to account for the \emph{social} aspects of learning. 
This approach was pioneered by Rotter \cite{rotter1954social} and further developed by Bandura \cite{bandura1977social}.
Bandura's social learning theory augments trial-and-error, stimulus-response behaviorism with more cognitively informed accounts of learning.
More specifically, Bandura posits that reinforcement may be \emph{vicarious}, i.e., obtained through a \emph{model} that is separate from the learning subject.
Models may be live, verbal or symbolic: 
a live model demonstrates some behavior by acting it out, whereas a verbal model gives an instruction or suggestion. Lastly, symbolic models may be fictional characters presented through print or some other media.
Beyond observation and blind imitation, learners in the social learning theory framework are afforded \emph{motivation}: the capability to observe the reinforcement signal obtained by the model and reason about whether they should actually imitate it.

In social psychology, considerable attention has been given to the phenomena of social facilitation and social inhibition, defined as the phenomenon of improved or worsened task performance in the presence of other individuals \cite{geen1991social,guerin1983social,zajonc1969social}.
Moreover, the effectiveness of collaborative learning as a complementary or alternative approach to teacher-centered instruction remains a subject of debate.
In this domain, perhaps the most well-known investigation is due to Whitman and Fife \cite{whitman1988peer}, who described and developed ``peer-to-peer teaching'' strategies in higher education; recent results, both survey-based~\cite{rybczynski2011student} and in controlled studies~\cite{jain2015impact}, have shown that the more competent agents rarely benefit from group collaboration.

Inspired by social learning theory and social psychology, we introduce \emph{peer learning} (see \cref{fig:peer_learning}), a learning scheme that enables knowledge exchange and social learning in groups of RL agents.
Our setup investigates the reward trajectories of agents which \emph{train together} but \emph{perform alone}, similar to students engaging in collaborative or cooperative learning, e.g., a study group, who might still have to pass an individual exam. This is a common behavior in human learning. 
Humans rarely learn a task alone and from scratch. 
Although infants start by trial-and-error learning, they will observe their parents \cite{observational_learning} and as they grow older, at some point, they will learn from experts such as teachers or coaches.
In contrast, for intelligent agents, the learning is often modelled through pure RL, hence, the agent is isolated in its learning setup \cite{Sutton2018}.
Consequently, similar to social learning theory, we introduce \emph{peer agents} as capable of learning either from pure stimulus-response or from \emph{vicarious reinforcement}.
We model vicarious reinforcement by allowing agents to observe other peers' suggestions for the situation (state) they are in.  
Different from ``blind'' imitation learning approaches, e.g., \cite{DAgger}, agents are afforded motivation and may decide whether to trust their peers' advice in the future. 
Motivation is modeled via a \emph{trust mechanism}, implying that at each learning step each agent needs to decide whether to follow the advice they have been given based on their previous experiences (and rewards) following the advice given by the peer/model.
We formalize the trust mechanism as a non-stationary, multi-armed bandit problem.
We find that groups of agents are able to isolate malicious, adversarial peers, but that honest, low-performing agents can still improve the performance of other learners.

Compared to previous agent-to-agent communication strategies (e.g., \cite{nunes2002,daSilva2017}), we are interested in cooperative, \emph{tabula rasa}
learning without expert advice---we do not assume the presence of a teacher who, by definition, gives trustworthy advice---, and our peers start out as complete novices in all the considered tasks. 
Furthermore, we test our framework's performance in learning motor skills via the MuJoCo control suite \cite{mujoco}, whereas agent-to-agent communication strategies have been so far limited to discrete action spaces \cite{cheng2021,daSilva2020,ilhan2021,omidshafiei2019}.
On this regard, our results stand in contrast to studies in social psychology \cite{bond1983social} and collaborative learning \cite{rybczynski2011student,jain2015impact}: the presence of multiple agents is beneficial both in learning complex motor skills and navigation in simpler discrete environments.

Our contribution may be summarized as follows:
\begin{itemize}
    \item[\textbullet] We introduce \emph{peer learning}, an \emph{action advice framework} for reinforcement learning (RL) in groups of agents (\emph{peers}).
    \item[\textbullet] It works for \emph{discrete} as well as \emph{continuous action spaces},
    \item[\textbullet] with almost every \emph{off-policy RL algorithm},
    \item[\textbullet] featuring a \emph{trust mechanism} that is able to identify the performance of peers within the group well
    to (i) boost performance over single agent training and the baseline
    and (ii) prevent poisoning attacks from malicious, adversarial agents.
\end{itemize}
%
%
\section{Related Work}
\label{sec:related_work}
Imitation learning and behavior cloning have a relatively long tradition in ML---especially in supervised learning but also RL---with the first approaches appearing in the 90s \cite{bain1995framework,ng2000algorithms}.
In these setups an agent learns from demonstration, which is provided by an expert---be it another software agent or a human.
This avenue of research has generated a surge of recent interest, with many approaches studying the problem of inverse RL~\cite{arora2021survey}, i.e., learning a reward function from examples.
Our approach differs in that we seek to model the social aspect of learning and allow for communication between concurrently learning agents, displaying motivation in the sense of social learning theory (see \cref{sec:intro}), which is closely related to what is referred to as trust in the advice exchange community.

In multi-agent RL (MARL) \cite{zhang2021multi}, several agents act in the same instance of an environment and need to compete or cooperate to solve the task at hand. 
While some recent approaches in MARL have investigated inter-agent communication \cite{zhu2022survey}, our approach focuses instead on separate instances of the same single-agent environment.
The objective of our investigation is to understand whether communication is intrinsically beneficial to the learning of complex, continuous-valued actions in single-agent learning.
Therefore, our setup is more akin to social learning theory and collaborative learning \cite{jain2015impact,rybczynski2011student}: We allow students to give each other action recommendations during the learning process, whereas the testing (inference) is done on an individual basis.

More closely related to this work are proposals for \emph{advice exchange} in RL.
The first approaches in this area are due to the work of Nunes and Oliveira \cite{nunes2002,nunes2003} and rely on tabular methods to learn strategies in grid-world environments.
In the advice exchange framework, as originally developed, learners act in a single multi-agent environment where the optimal policy requires collaboration at testing time.
Advice is defined as the action the adviser would perform if it found itself in the same state as the advisee---from now on called \emph{action advice}.
While some algorithmic proposal to learn which agents give trustworthy advice is already present in this early work~\cite{nunes2002}, the authors do not consider scenarios in which adversarial, bad-faith agents may be part of the study group.
Notably, the recommendation to perform further work in the space of motivation and trust is also present in more recent work in advice exchange \cite{bignold2021,omidshafiei2019,daniluk2020,daSilva2020}, but is actually undertaken only in a minority of methods \cite{ndousse2021,cheng2020}. 
Another limitation of existing approaches in advice exchange is that they might require fully-trained teacher agents---also called experts or oracles---, either during training \cite{daSilva2020uncertainty,daSilva2017,cheng2020,ilhan2021} or even at test time \cite{ndousse2021}.
Finally, to the best of our knowledge, previous advice exchange proposals have focused on discrete action spaces and relatively simple navigation-based skills.
In contrast, our approach is i) to design motivation and trust mechanisms to be employed by ii) non-expert agents which learn concurrently, starting from scratch iii) on continuous action spaces representing fine-grained motor skills.
In the interest of enabling a comparison to work performed in advice exchange, we compare our methodology to the more recent methods which do not explicitly require fully-trained agents \cite{omidshafiei2019,torrey2013}.
%
%
\section{Peer Learning}
\label{sec:method}
\textbf{Motivation.} Peer learning introduces a training paradigm rooted in pure reinforcement learning (RL) while also incorporating social learning principles.
Our framework, represented in \cref{fig:peer_learning}, is analogous to a study group where numerous peers undertake simultaneous training. 
Each agent engages with \emph{its own} environment. Even though all environments mirror each other, they exist independently, each embodying the same Markov Decision Process (MDP).
Agents may exhibit social behavior by sharing action recommendations, which allows the learning process to be analyzed through the lens of social learning theory \cite{bandura1977social}.
In peer learning, each agent operates independently: Peers do not operate cooperatively, and environments remain detached, contrasting with Multi-Agent RL \cite{omidshafiei2019} and prior advice exchange proposals \cite{ndousse2021}.
The rationale behind exploring this arrangement is to discern whether independent agents exhibit vicarious reinforcement potential---the ability to evaluate peers' actions and determine whether it would be beneficial to imitate them.

In practical terms, we suggest that this framework is suitable for shared, cross-institutional training of RL agents.
Under a ``peer learning'' configuration, various autonomous organizations would train their RL agents to execute an identical task, such as exploration in adverse conditions or autonomous driving.
In this context, details regarding agent design (action space, reward) and world representation (state space) could feasibly be negotiated in advance; nevertheless, the task should continue to be performed independently following successful training.

\textbf{Definition.} Peer learning can be characterized by 
the tuple $\mathcal{P} = (\mathcal{N},{\mathcal{M}_i})$, in which $\mathcal{N}$ signifies a set of agents ${0, \dots, n-1}$ and ${\mathcal{M}_i}$ symbolizes a collection of distinct yet equivalent MDPs.
Each MDP is represented as a tuple $\mathcal{M} = (\mathcal{S}, \mathcal{A}, \mathcal{R}, \mathcal{T}, \mu, \gamma)$.
In the context of this MDP, at every step $t$, the agent surveys the state $S_t \in \mathcal{S}$, with $S_0$ drawn from the initial state distribution $\mu$, and undertakes the action $A_t \in \mathcal{A}$ in line with its policy $\pi_i(A_t | S_t)$. Subsequently, the environment transitions from state $S_t$ to $S_{t+1} \sim \mathcal{T}(S_t, A_t) = \mathrm{Pr}(S_{t+1}|S_t, A_t)$ and produces the immediate reward $R_{t+1} = \mathcal{R}(S_t, A_t, S_{t+1}) \in \bbbr$, which is observed by the agent.
Peer learning can then be defined as the concurrent training of $n = |\mathcal{N}|$ RL agents, all participating in identical versions of a singular MDP $\mathcal{M}$.
Every agent $i \in \mathcal{N}$ aims to amplify its own discounted long-term reward $G_{t,i}$:
\begin{equation}
G_{t,i} = \mathbb{E}_{\pi_i} \left[ \sum_{k=t+1}^{T}\gamma^{k-t-1} R_{k,i} \right],
\end{equation}
where $\gamma \in [0, 1)$ denotes the discount factor. This return can be estimated by the value function $V_{\pi_i}(S_{t,i})$, computing the anticipated return from state $S_{t,i}$ while adhering to policy $\pi_i$. Furthermore, we can define the Q-function $Q_{\pi_i}(S_{t,i}, A_{t,i})$ as the expected return following the action $A_{t,i}$ in state $S_{t,i}$ and implementing the policy $\pi_i$ thereafter.

While this formalization is not the sole way to define peer learning, it does offer a direct and intuitive interpretation of the collaborative learning task at hand.
An alternative definition, which we only sketch in the following due to space constraints, is to characterize peer learning as a decentralized MDP, which is additionally state-factored, transition-independent, and reward-independent. 
We refer the reader to the seminal work by Becker et al. \cite{becker2003transition} for formal definitions of these properties. 

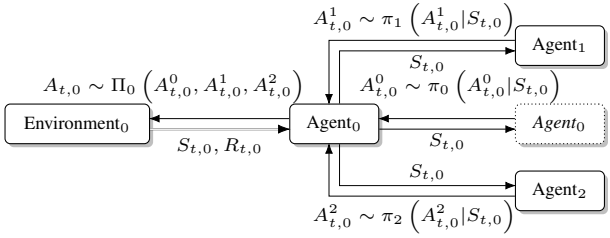
\begin{figure}[t]
\centering
\begin{tikzpicture}[font=\scriptsize]
    \newlength{\arrowdist}
    \setlength{\arrowdist}{0.7mm}
    \node[frame] (agent0) {Agent$_0$};

    \node[densely dotted,frame,xshift=3cm] (agent0_) {\textit{Agent$_0$}};
    
    \node[frame,above=5mm of agent0_] (agent1) {Agent$_1$};
    \node[frame,below=3.5mm of agent0_] (agent2) {Agent$_2$};
    
    \node[big_frame,left=1.85cm of agent0] (env) {Environment$_0$};
    
    \draw[line,] ([yshift=\arrowdist]agent0.west) -- ([yshift=\arrowdist]env.east) node[pos=0.83,above=0.9mm] {$A_{t,0}  \sim \Pi_0\left(A_{t,0}^0, A_{t,0}^1, A_{t,0}^2\right)$};
    \draw[line,double] ([yshift=-\arrowdist]env.east) -- ([yshift=-\arrowdist]agent0.west) node[midway,below] {$S_{t,0},R_{t,0}$};
    
    \draw[line] ([yshift=\arrowdist]agent0_.west) -- ([yshift=\arrowdist]agent0.east) node[pos=0.4,above=0.9mm] {$A_{t,0}^0 \sim \pi_0\left(A_{t,0}^0|S_{t,0}\right)$};
    \draw[line] ([yshift=-\arrowdist]agent0.east) -- ([yshift=-\arrowdist]agent0_.west) node[midway,below=-0.6mm] {$S_{t,0}$};
    
    \draw[line] ([xshift=\arrowdist]agent0.north) |- ([yshift=-\arrowdist]agent1.west) node[near end,below=-0.6mm] {$S_{t,0}$};
    \draw[line] ([yshift=\arrowdist]agent1.west) -| ([xshift=-\arrowdist]agent0.north) node[pos=0.27,above=0.5mm,inner sep=0] {$A_{t,0}^1 \sim \pi_1\left(A_{t,0}^1|S_{t,0}\right)$};
    
    \draw[line] ([xshift=\arrowdist]agent0.south) |- ([yshift=\arrowdist]agent2.west) node[near end,above=-0.7mm] {$S_{t,0}$};
    \draw[line] ([yshift=-\arrowdist]agent2.west) -| ([xshift=-\arrowdist]agent0.south) node[pos=0.27,below=0.5mm,inner sep=0] {$A_{t,0}^2 \sim \pi_2\left(A_{t,0}^2|S_{t,0}\right)$};
\end{tikzpicture}
\caption{An intricate depiction of the advice exchange in peer learning with three agents. Agent $0$ operates within its environment and requests advice in state $S_{t,0}$ from all peers in the group, including itself (outlined in dotted frame). In return, it receives suggested actions $A_{t, 0}^0, \dots, A_{t, 0}^2$ and then selects an action from those suggestions according to its peer policy $\BigPi_0$.}
\label{fig:peer_learning_details}
\end{figure}

\textbf{Agent communication.} The $n$ agents in a peer learning setup are able to communicate in a constrained manner. Each agent $i \in \mathcal{N}$ solicits action advice $A_{t, i}^j \sim \pi_j(A_{t, i}^j | S_{t, i})$ from all other agents $j$ ($j \in \mathcal{N}$) based on the state $S_{t,i}$ it currently occupies and the policies of all agents. Agents then decide from the compilation of all proposed actions: $A^0_{t,i}, \dots, A^{n-1}_{t,i}$, including their own, in accordance with a distinct \emph{peer} policy $\BigPi_i(A_{t,i}^0,\dots,A_{t,i}^{n-1})$. Therefore, in peer learning, each agent has the ability to produce actions---and action recommendations---through its own policy $\pi_i$ and select the most beneficial one through the peer policy $\BigPi_i$. A detailed depiction of peer learning with three agents is presented in \cref{fig:peer_learning_details}.

While the primary goal of peer learning still aligns with the conventional RL aim of maximizing the discounted long-term reward of the MDP, we can also frame a similar goal for the sub-task of requesting and choosing action advice from peers. Here, the objective is for each agent to learn about its peer's proficiency level, that is, estimating the quality of their advice. This task is essentially a \emph{multi-armed bandit problem} \cite{banditsintro}
, where the goal is to maximize the reward and thus identifying the ``best'' bandit. Similarly, in peer learning, each agent needs to identify the most proficient peer from whom to seek advice. Unlike the standard multi-armed bandit problem, in peer learning, we are dealing with a non-stationary multi-armed bandit problem. Therefore, more recent rewards should be given more weight than long-past ones \cite{Sutton2018}.

Framing the problem of taking advice in this manner underscores the \emph{need for exploration} in terms of whom to ask for advice or whose advice to heed, respectively. Surprisingly, this critical aspect is often overlooked in most previous work and has recently been mentioned as one critical avenue for further research in advice exchange techniques \cite{daSilva2020}. Consequently, we propose our approach as follows.

During the training process, agent $i$ must decide on one action recommendation $A_{t,i}^j$ to be actually performed. This decision is made via weighted sampling with weights $v_i^j$ using a Boltzmann (or Gibbs) distribution
\begin{align}
A_{t,i} &\sim \BigPi_i(A_{t,i}^{0},\dots,A_{t,i}^{n-1})\nonumber\\
&= \mathrm{Pr}(A_{t,i} = A_{t,i}^j | S_{t,i})
= \frac{e^{v_i^j / \temp_{\epoch} } }{ \sum_k e^{v_i^k / \temp_{\epoch}} }, \label{eq:vij} \\
&\text{with } A_{t,i}^j \sim \pi_j\left(A_{t,i}^j | S_{t,i}\right), \nonumber\\
&\text{and }\temp_{\epoch} = \tau_{0} \; e^{- \lambda\epoch}, \nonumber
\end{align}
where $i,j,k \in \mathcal{N}$, $\epoch$ represents the current epoch, $\temp$ is the temperature parameter, and $\lambda$ is the temperature decay. For $\temp_0 = 1$ and $\lambda = 0$, the Boltzmann distribution\footnote{We replace the negative energy $-E$, that is commonly used---especially in physics---, by our positive weights $v$.} equates to the softmax function. Hence, we can also interpret it as a weighted softmax function.
The weight $v^j_i$ has a critical function here, as it represents the level of trust, or motivation, that agent $i$ has in its peer $j$.
As previously mentioned, our framework makes it possible to learn this weight adaptively at training time. 
In the following, we formalize three different techniques to do so.

\textbf{Critic: Advice Evaluation via Q-function}. The simplest evaluation technique we propose is to employ agent $i$'s Q-function to evaluate the \emph{advice} received, independently of which agent $j$ has sent it. Specifically, for each agent $i$ and every other agent $j$ the advice weight is defined as
\begin{equation}
    v_i^j = Q_i(S_{t, i}, A_{t, i}^j)
\end{equation}
We dub this approach \textbf{Critic} as the Q-function is commonly referred to as the critic in actor-critic RL algorithms.
One limitation of this approach is that $Q_i$ could fail in recognizing valuable advice. 
This is especially problematic at the beginning of training---agents could reject advice from an expert simply because their own Q-function is currently underfit.
Furthermore, this technique only evaluates the advice, and not the advice-giver.
Thus, it could be difficult for the peer group to isolate bad-faith actors who participate in the training with adversarial objectives. 
In the following, we propose two other motivation mechanisms that seek to solve these issues.

\textbf{Trust Values: Local Evaluation of Peers}. In this mechanism, we seek to compute $v_i^j$ based on the quality of the advice that agent $j$ gave to its peer $i$ throughout the training process.
At a basic level, this may be computed by simply keeping track of the immediate rewards incurred by $i$ when following advice given by $j$.
Starting from this straightforward concept, we extend the experience replay buffer \cite{dqn2} so that peers may keep track of whose advice they previously followed and how beneficial it was. 
Specifically, experience replay is usually defined as a tuple $<S_t; A_t; R_{t+1}; S_{t+1}>$.
In this work, we propose to extend it to the tuple $<S_{t, i};A_{t,i}^j;R_{t+1,i}^j;S_{t+1,i}^j;j>$ where, with a slight abuse of notation, we use the variable $R_{t+1,i}^j$ to denote the immediate reward that agent $i$ obtained by following the advice of agent $j$ at step $t+1$. 
We then initialize $v_i^j$ randomly and update it throughout training by bootstrapping $R_{t+1,i}^j$ with agent $i$'s Q-function, as follows:
\begin{align}
    v_i^j &\leftarrow (1 - \alpha) v_i^j + \alpha \omega^j_i \\
    \omega^j_i &= R_{t+1, i}^j + \gamma Q_i(S_{t+1, i}^j, \pi_i(S_{t+1, i}^j)) \label{eq:omega},
\end{align}
where $\alpha$ is a ``trust learning rate'' separate from the base learning algorithm's learning rate. In our experiments, we keep this fixed at $0.99$.

\textbf{Agent Values: Global Leaderboard of Peers}. To introduce this last motivation technique, we start from the observation that sharing information between peers about who is giving good advice may be beneficial to quickly learn to identify experts or bad-faith actors.
To this end, we introduce the idea of maintaining a global experience buffer ${<}S_{t, i};A_{t,i}^j;R_{t+1,i}^j;S_{t+1,i}^j;j{>}$ and computing a global weight $v^j$ to be substituted in for $v_i^j$ in \cref{eq:vij} for all $i$:
\begin{align}
    v^j &:= v_0^j = v_1^j = \dots = v_{n-1}^j \\
    v^j &\leftarrow (1 - \alpha) v^j + \alpha \omega_i^j, \quad \forall i \in \mathcal{N}
\end{align}
\textbf{Learning the Advantage of Following Advice}. While the techniques introduced so far all are able to effectively model the motivation of trust that each agent has in following the advice given by other peers, we find that it is beneficial for the agents to compare the benefit of participating in peer learning with the reward that they would obtain by learning independently.
As a result, we also train agents which update their trust and agent values \emph{with advantage}, that is, by computing $v_i^j$ as the difference between the quality of the advice received and the action they themselves would have taken if isolated. 
In practice, this means changing \cref{eq:omega} to
\begin{equation}
    {%
    \omega^j_i = R_{t+1, i}^j \!+\! \gamma Q_i(S_{t+1, i}^j, \pi_i(S_{t+1, i}^j)) \!-\! Q_i(S_{t, i}, \pi_i(S_{t, i})),
    \label{eq:advantage}}
\end{equation}
where the difference to \cref{eq:omega} is in the last subtracted term.
%
%
\section{Experiments}
\label{sec:experiments}
Our experimentation seeks to understand the capabilities of peer learning in two different contexts: \emph{learning complex policies}, i.e., in the continuous state and action spaces provided by the MuJoCo control suite; furthermore, we elaborate on the effect of \emph{motivation}, or trust mechanisms, introduced in the previous section. 
Our findings can be summarized as follows:
\begin{itemize}
    \item[\textbullet] Vicarious reinforcement via peer learning is able to improve performance in the MuJoCo control suite when compared with single-agent learning. We discuss these results further in \cref{ssec:comparison}.
    To the best of our knowledge, our contribution marks the first time in which techniques based on the concept of exchanging advice have been proven successful in such environments.
    \item[\textbullet] In \cref{ssec:ablation_study}, we show how the motivation mechanisms introduced in \cref{sec:method} may be employed to find and ``isolate'' bad-faith peers. 
    We investigate this effect in a room-like environment where agents need to navigate to a certain position to obtain the reward.
    We find that our agents are able to ignore advice coming from non-trustworthy peers, even when the short-term rewards for following bad advice are not worse.
    This situation has been previously called ``poisoning attack'' in the literature~\cite{cheng2021}.
\end{itemize}

\noindent We consider the following learning algorithms and agents:
\begin{description}
    \item[Peer Learning agent]
    A learning agent in the newly introduced peer learning setting with advantage which gives and receives advice through action recommendations for specific environment states within its peer group.

    \item[Single agent]
    A normal reinforcement learning single agent which trains alone in an MDP without advice exchange.
    
    \item[LeCTR agent \cite{omidshafiei2019}]
    A state-of-the-art action advising agent which learns a task-level execution policy and a teaching policy for action advising its peer.
    
    \item[Early Advising agent] An agent which only exchanges action advice in the early phase of training until the teaching budget exhausts. Like, e.g., used by  \cite{omidshafiei2019} and \cite{torrey2013}.
    
    \item[Adversarial agent] An adversarial agent as defined by \cite{cheng2021} with the objective to minimize its peers' long-term reward without changing the immediate reward.
    
    \item[Non-training novice] A randomly initialized agent which does not update its policy and/or Q-function.
    
    \item[Peer Learning (Random advice)] An agent similar to the peer learning agent but choosing randomly among all the suggested actions.
\end{description}
\subsection{Implementation}
\label{ssec:implementation}
Our Python code can be found on GitHub%
\footnote{\url{https://github.com/kramerlab/PeerLearning}}
and works with several off-policy RL algorithms that make use of a Q-function---especially but not limited to actor-critic methods.
For most of our experiments, i.e., in the continuous action space settings, we used the state-of-the-art Soft-Actor Critic (SAC) algorithm \cite{SAC} as basis for our setup. 
For our experiments with discrete action spaces, we chose the widely used vanilla Deep Q-networks (DQN)\cite{dqn1,dqn2} algorithm.
\if T\arxiv@on%
Further information on our code and reproducibility can be found in \cref{sec:hardware}.%
\else%
Further information on our code and reproducibility can be found in the appendix of the ArXiv version of this paper \cite{peerlearning_arxiv}.
\fi

\begin{figure}[t]
    \centering
    \resizebox{!}{0.27\columnwidth}{
    \begin{tikzpicture}
        \draw [step=1.0,black, thick] (0,0) grid (5,5);
        \node[] at (2.5,2.5) {\includegraphics[width=0.9cm]{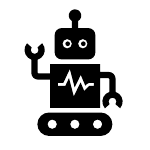}};
        \node[] at (4.5,0.5) {\includegraphics[width=0.5cm]{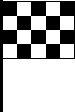}};
    \end{tikzpicture}
    }
    \caption{Our {\tt{}Room} environment of size $5 \times 5$. The agent (robot) starts in the center of a square grid-world, where the goal position is a random position at the border (flag). It can choose among the four actions: right, left, up and down.}
    \label{fig:room}
\end{figure}
\begin{table*}[ht]
    \fontsize{9pt}{9pt}\selectfont
    \centering
    \begin{tabular}{c S[table-format=4.0(3)] S[table-format=4.0(3)] S[table-format=+4.0(3)] S[table-format=4.0(3)] S[table-format=2.0(1)]}
        \toprule
   {Task}          & {Peer Learning}  & {RL Single Agent} & {\;\;Early Advice} & {Peer Learning (Random Advice)} & {LeCTR}  \\
    \midrule
    HalfCheetah-v4 & \bfseries 9014(715)            & \bfseries 9270(247)  & 7553(1284)           & \bfseries 9129(504)  &  \text{-}                 \\
    Walker2d-v4    & \bfseries 2160(346)            & 1558(489)            & 248(245)             & 1970(526)            &  \text{-}                \\
    Ant-v4         & 2459(667)                      & \bfseries 2694(660)  & -2222(114)           & 2109(632)            &  \text{-}                 \\
    Hopper-v4      & 2123(963)                      & 2059(666)            & 549(381)             & \bfseries 2483(87)   &  \text{-}                  \\
    Room-v21       & \bfseries 72(8)                & 46(16)               & 12(4)                & 68(10)               & 15(3)                \\
    Room-v27       & 58(11)                         & 32(13)               & 8(5)                 & \bfseries 69(11)     & 5(2)                 \\
    \bottomrule
    \end{tabular}
    \caption{Comparison of learning speed and final performance expressed as average reward over time ($\pm$ the standard deviation). Values within 5 percent of the maximum are printed bold. For all experiments, we used 10 random seeds except for the Room-v27 where we used 15.}
    \label{tab:comparison}
\end{table*}
Being among the small group of approaches that work for continuous action spaces, we used several MuJoCo \cite{mujoco} OpenAI Gym environments \cite{gym}, i.e., {\tt{}HalfCheetah-v4}, {\tt{}Walker2d-v4}, {\tt{}Ant-v4} and {\tt{}Hopper-v4}. 
Apart from the MuJoCo environments, for comparison to our baseline and poisoning attacks \cite{cheng2021}, we also use our room grid-world environment displayed in \cref{fig:room}. In this environment, the agents starts in the middle of a squared grid and its task is to reach a specific goal position, which is chosen randomly among the border position. The agent observes the position of the goal next to its own position.
The exploration in this task is more complex than in the usual grid-world tasks and satisfies all requirements for our baseline.
In \cref{fig:room}, we display the environment with a small grid size of $5 \times 5$ for visualization purposes. In our experiments, we use a bigger grid sizes of 
$27 \times 27$ and $21 \times 21$ creating the right amount of challenge for the agent.

\subsection{Learning Complex Policies from Scratch}
\label{ssec:comparison}
We test our approach in the MuJoCo control suite, comparing it against independent agents without communication and the commonly employed baseline Early Advising \cite{torrey2013}. 
In this setting, we use a group size of 4 peers\footnote{Although we only use a fixed number of agents during our experiments, in theory peer learning can deal with a changing number of peers when the maximum number of peers is known beforehand.}.
These results can be found in \cref{tab:comparison}, which displays the average reward over the learning process---which can be seen as a more interpretable equivalent to the area under the learning curve---as a measure of learning speed.
We observe that peer learning (with advantage) gives stronger results compared to the baselines in 3 out of 4 environments.

In \texttt{Ant-v4}, employing peer learning accelerates training: however, the benefits of agent communication seem to decrease over time.
We conclude that vicarious reinforcement is a strong learning strategy in RL when learning complex motor skills, in contrast to early results on the topic \cite{bond1983social}.
\subsection{The Role of Trust and Motivation}
\label{ssec:ablation_study}
\begin{figure}[t]
    \centering
    \includegraphics[page=2]{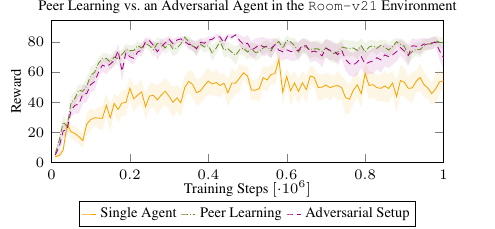}
    \caption{The reliability mechanisms in our approach protect the learning against the influence of a malicious agent. Peer learning cannot be harmed by an adversarial agent and still outperforms single agent learning.}
    \label{fig:adversarial}
\end{figure}%
As discussed in \cref{sec:method}, we proposed different mechanisms to deal with the motivation/trust problem---i.e., deciding whether to follow advice and which advice is most promising. 
To verify the positive effect of each of our proposed mechanisms and to find a best combination, we show the results of our combination study in \cref{tab:ablation_study}.
We have averaged results over multiple MuJoCo environments and normalized them between 0 and 100. We also use the average 
normalized cross-environment reward over time as measure for the learning speed.
We observe any of our proposed methods boosts performance over single agent learning. 
When using a combination of at least two of our proposals, the performance increases again by a small margin. 
Nonetheless, our study did not yield a single best combination that significantly outperforms all the other combinations in all environments. We can, however, notice that a combination of all three ranks in the leading group, especially when using advantage.
Hence, we used advantage and an equally weighted combination\footnote{For equal weighting, we normalize each term to $[-1, +1]$.} of all three approaches, i.e., critic, trust and agent values, for our further experiments.%
\begin{figure}[t]
    \centering
    \includegraphics[page=5]{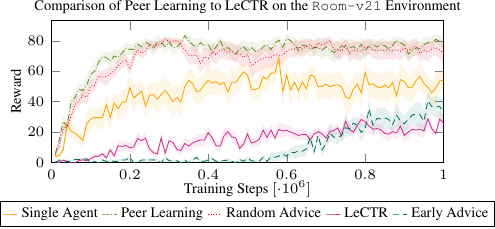}
    \caption{In a comparison, our approach (\emph{Peer Learning}) outperforms the \emph{LeCTR} approach \cite{omidshafiei2019} and a solo RL agent (\emph{Single Agent}) on average over 10 random seeds on the Room environment of size $21 \times 21$. The shaded area marks the standard error of the mean.}
    \label{fig:comparison_room}
\end{figure}%
\begin{figure*}[ht]
    \centering
    \includegraphics[page=6]{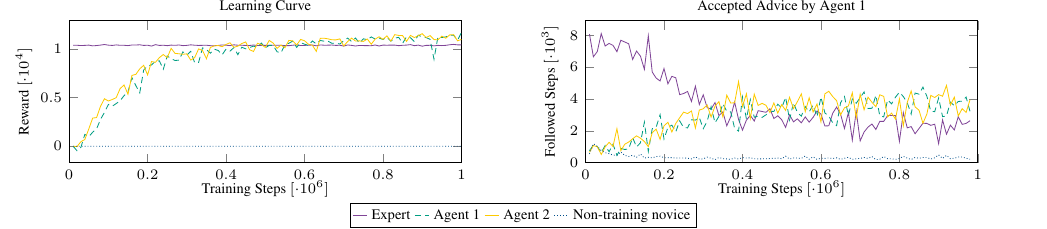}
    \caption{In a study group with one domain expert and a non-training novice, the two peer learning agents with advantage learn very well which of its peers gives good advice and whom to avoid. They also learn to switch to another peer if the best performing agent gets overtaken.}
    \label{fig:expert}
\end{figure*}%

When employing this setup, we find that peers are able to identify bad actors and ignore their advice: 
We train a group of peers on a grid-like environment (see \cref{fig:room}) where agents only obtain rewards when they position themselves in a certain randomly sampled location at the borders of the rooms, a ``goal position''. 
In this setting, we include one adversarial agent which always gives out the suggestion for another agent to distance itself from the goal.
We note that the short-term reward for following this advice is aligned with the short-term reward for following the theoretically optimal advice, as rewards are only obtained when reaching the goal position exactly; this follows the design of a ``poisoning attack'' for advice exchange as proposed by Cheng et al.~\cite{cheng2021}.
Results for this experiment can be found in \cref{fig:adversarial}.
Here, it is apparent that employing our suggestion selection mechanisms increases performance substantially.
More generally, this simpler room environment enables us for a broader comparison with other methods in advice exchange which only support discrete state, and more importantly, action spaces.
We limit our comparison to methods which do not explicitly require pre-trained experts: LeCTR~\cite{omidshafiei2019} and Early Advising~\cite{torrey2013}.
It is discernible that our methodology is able to outperform these two methodologies in the environment we considered, even over multiple room dimensions (\cref{fig:comparison_room}, \cref{tab:comparison}).
We also see that training the suggestion selection mechanisms with advantage (see \cref{eq:advantage}) allows peers to identify experts: \Cref{fig:expert} shows the number of times the learning agents accepted advice from its peers side-to-side with the learning curve.
The group of learners here was formed by 2 training novices, 1 non-training ``expert'' and 1 non-training novice. 
We obtain non-training agents by straightforwardly setting their learning rate to 0: a non-training novice will give arbitrarily bad advice throughout training.
\Cref{fig:expert} shows that the 2 training novices, i.e., peers, are able to identify that one agent is giving out particularly bad advice. It also emphasizes that they are quick to recognize when the ``expert'' is being overtaken.%
\begin{table}[t]
    {\setlength\tabcolsep{2pt}%
    \fontsize{9pt}{9pt}\selectfont%
    \centering
    \begin{tabularx}{\columnwidth}{YYYYYYYYYc}
    \toprule
    Adv        & ACT            &  AC            & AT             &  CT            & C                     &  A             &  T              &  RL                      & Random                   \\
    \midrule
    No         &           66.2 &           64.9 &  \textbf{68.2} &           65.7 & \multirow{2}{*}{65.4} &           66.4 &           66.4  & \multirow{2}{*}{59.7} & \multirow{2}{*}{65.0} \\ 
    Yes        &  \textbf{67.1} &  \textbf{67.1} &           65.8 &           64.7 &                       &           64.3 &            62.6 &                       &                       \\
    \bottomrule
    \end{tabularx}}
    \caption{Averaged over the 4 MuJoCo environments, it is evident that peer learning of any kind boosts training significantly compared to the single agent (RL), slightly against the peer learning with random advice agent (Random) and that several combinations of the agent values (A), critic (C) and trust (T) lead to great performance. The maxima are printed bold. For this ablation study, we used 5 random seeds.}
    \label{tab:ablation_study}
\end{table}%
\begin{figure}[t]
    \centering
    \includegraphics[page=1]{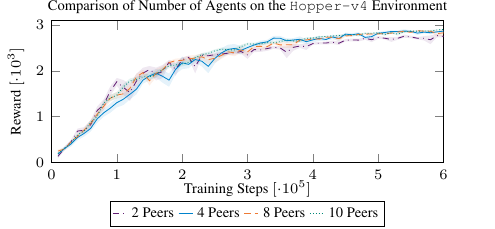}
    \caption{In peer learning, the performance slightly increases with the number of agents in the setting and does not peak or saturate for 2 agents. Averaged over 10 random seeds.}
    \label{fig:num_agents}
\end{figure}%
\subsection{Number of Agents}
\label{ssec:num_agents}
The literature suggests that a setup with 2 or 3 agents or experts, respectively, yields the best results, concluding that the overhead of learning which advice to follow hinders learning when the group becomes too big  \cite{cheng2020,kretchmar2003}.
In \cref{fig:num_agents}, we show that our results do not align with that hypothesis and that in peer learning the performance increases with the number of agents. Although the increase is not significant and further experiments on the limit have to be pursued, it is clearly visible that only 2 simultaneously learning agents do not outperform bigger groups of size 4 to 10.
%
%
\section{Conclusion and Future Work}
\label{sec:conclusion}
We have introduced a novel high-level framework called peer learning that enables social learning in groups and is compatible with many off-policy RL algorithms.
In our experiments, we have shown that the introduced peer learning framework was able to boost RL and outperform a single agent (\cref{ssec:comparison}). 
Incorporating trust mechanisms that take into account the quality of other agents allows for a stronger focus on the good peers in the group.
Our setup is therefore able to avoid poisoning attacks from adversarial agents. 
Furthermore, peers whose skills are comparatively more developed than the rest of their group may also rely on motivation and trust to reject untrustworthy advice (\cref{ssec:ablation_study}).
To the best of our knowledge, we are the first to identify this trust problem as a non-stationary multi-armed bandit problem, and to successfully employ such a mechanism in learning complex continuous policies.
Our current proposal, while valuable, does not define a singular method for implementing trust and motivation. As a result, it falls short of achieving a comprehensive understanding of social learning theory in RL.
The evidence we gathered suggests that, while employing trust can be beneficial (\cref{tab:ablation_study}) and is critical when not every peer is a trusted partner (\cref{fig:adversarial}), there is no single superior choice among our three proposals.
This warrants future investigation, especially in setups with more than 10 concurrently learning agents (\cref{ssec:num_agents}).
While our conceptualization of peer learning is embarrassingly parallel, our current implementation is not. As a consequence, this work does not discuss the properties of peer learning in setups with, e.g., tens of agents and leaves it to future work.
%
%
\section*{Ethical Statement}
No known ethical concern is implied by the presented work.
%
%
\section*{Acknowledgements}
This research project was partly funded by the Hessian Ministry of Science and the Arts (HMWK) within the projects ``The Third Wave of Artificial Intelligence - 3AI'' and hessian.AI. This work was partly funded by the RMU by means of the DeCoDeML project. 
Parts of this research were conducted using the supercomputer MOGON 2 and/or advisory services offered by Johannes Gutenberg-University Mainz (hpc.uni-mainz.de), which is a member of the AHRP (Alliance for High Performance Computing in Rhineland-Palatinate, www.ahrp.info) and the Gauss Alliance e.V.
The authors gratefully acknowledge the computing time granted on the supercomputer MOGON 2 at Johannes Gutenberg-University Mainz (hpc.uni-mainz.de).
We further acknowledge the use of DALL·E 2 (Open AI, \url{https://openai.com/dall-e-2}) to create parts of \cref{fig:peer_learning}.
%

%
%
\bibliography{camera_ready}
%
%
\if T\arxiv@on
\clearpage
\appendix
\section{Hardware and Software Specifications}
\label{sec:hardware}
To conduct our experiments, we used the \textit{MOGON 2} high-performance computing cluster which comes with the CentOS Linux operating system and CPUs of type Intel Xeon E5-2630v4 (Broadwell architecture) and Intel Xeon Gold 6130 (Skylake architecture)\footnote{For further information, please see \url{https://hpc-en.uni-mainz.de/high-performance-computing/systeme/#Mogon_2-Cluster}.}.
All our experiments ran purely on one CPU and had up to 10GB of RAM available.

We attach great importance to understandable and usable source code and are advocates of open science. Hence, our commented Python code can be found on GitHub\footnote{\url{https://github.com/kramerlab/PeerLearning}}.
The software backbone of our implementation is the Stable-Baselines3 \cite{SB3} framework which is build upon PyTorch \cite{torch} and OpenAI Gym \cite{gym}. The respecting versions and further libraries necessary to run our source code are specified in the \texttt{requirements.txt} file. To verify our strong independence of underlying RL algorithm, we also included the Twin Delayed DDPG \cite{td3} algorithm as an option within our implementation.
In Addition, there is a detailed description on how to replicate our experiments and how to install the software, including all necessary hyperparameters.
\fi

\end{document}